# Selection Mammogram Texture Descriptors Based on Statistics Properties Backpropagation Structure

Shofwatul 'Uyun[1]
[1]Doctoral Student of Computer Science Gadjah Mada University, Yogyakarta, Indonesia
shofwatul.uyun@webmail.ugm.ac.id

Sri Hartati[2], Agus Harjoko[2], Subanar[3]
[2]Department of Computer Science and Electronics,
[3]Department of Mathematics,
[2,3]Faculty of Mathematics and Natural Sciences,
Gadjah Mada University, Yogyakarta, Indonesia
{shartati, aharjoko}@ugm.ac.id, subanar@yahoo.com

*Abstract*— Computer Aided Diagnosis (CAD) system has been developed for the early detection of breast cancer, one of the most deadly cancer for women. The benign of mammogram has different texture from malignant. There are fifty mammogram images used in this work which are divided for training and testing. Therefore, the selection of the right texture to determine the level of accuracy of CAD system is important. The first and second order statistics are the texture feature extraction methods which can be used on a mammogram. This work classifies texture descriptor into nine groups where the extraction of features is classified using backpropagation learning with two types of multi-layer perceptron (MLP). The best texture descriptor as selected when the value of regression 1 appears in both the MLP-1 and the MLP-2 with the number of epoches less than 1000. The results of testing show that the best selected texture descriptor is the second order (combination) using all direction ($0^0, 45^0, 90^0, 135^0$) that have twenty four descriptors.

I. INTRODUCTION

Number of cancer patients in the world increasing every year is 6.25 million people from developing countries including Indonesia. In Indonesia, breast and cervical cancers rank the highest in turn. Therefore, Indonesian women are expected to be more vigilant and continue making early detection to prevent this disease. For that reason, early detection is an important effort to prevent it [1]. Basically, there are two medical treatments for breast cancer, they are screening and diagnostics. Computer technology used for screening is commonly called Computer Aided Diagnosis (CAD) system, that is the most effective method to reduce the number of death caused by breast cancer. Many image format used for screening, the most widely used is mammogram [2] and [3]. Other work [4] has been done using ultrasound for breast cancer. CAD systems for mammogram has been much developed by previous researchers who have focused on the preprocessing, feature extraction and classification. They have used the MIAS and DDSM public database. The database have been classified and analyzed by the radiologist. GLCM has some parameters, Shesadri uses seven parameters of GLCM (mean, standar deviation, smoothness, third moment, uniformity and entropy). The results of the extraction with seven parameters are classified into four categories i.e. fatty, uncompressed fatty, dense and high. Thereafter, classification results are compared to the assessment by the radiologist with 78% accuracy [5]. While [6] using only three parameters of GLCM i.e. contrast, correlation and entropy, it is then classified using naïve bayes classifier whose accuracy of 82,40%. Maitra *et al* [7] also used the method of GLCM for extraction of texture with four parameters (contrast, entropy, homogeneity and correlation) with value d=1 pixel using four directions ($0^0, 45^0, 90^0, 135^0$) and compared that to each direction with two categories i.e. mass and nonmass. Martins et al [8] use texture and shape features of mammogram. Four texture descriptors have been used were contrast, entropy, energy and inverse difference moment using four directions ($0^0, 45^0, 90^0, 135^0$) and three distances (d=1,2 and 3). So, the overall descriptors were 48 texture descriptors = 4 direction x 3 distances x 4 descriptors. While the shape descriptors were eccentricity, circularity and convexity.

Some researches show that the better detection rate can be achieved by appropiate feature selection that must included in the system that may require the number of features. However, having more features increases the complexity and time used to analyze the digital mammogram. In this paper, a comparison of first order and second order statistic texture descriptors is describe and the result are use for input classification . The classification using two types of backpropagation neural network.

II. THE PROPOSED MODEL

A method proposed for the development of CAD system consists of three stages : pre-processing, feature extraction and classification, which is shown in figure 1.

*A. Materials*

The data used in this work was taken from a public database MIAS (*Mammography Image Analysis Society*). MIAS [15] consists of 322 images of 161 patients with MLO view (Mediolateral Oblique), which is the result of digitizing scanner with a resolution of 50 microns and the PGM (portable graymap format) with a size of 1024x1024. The MIAS data was classified and validated by the radiologist into benign (54 images) and malignant (39 images). The fifty cases were selected randomly from a total of 93 images.

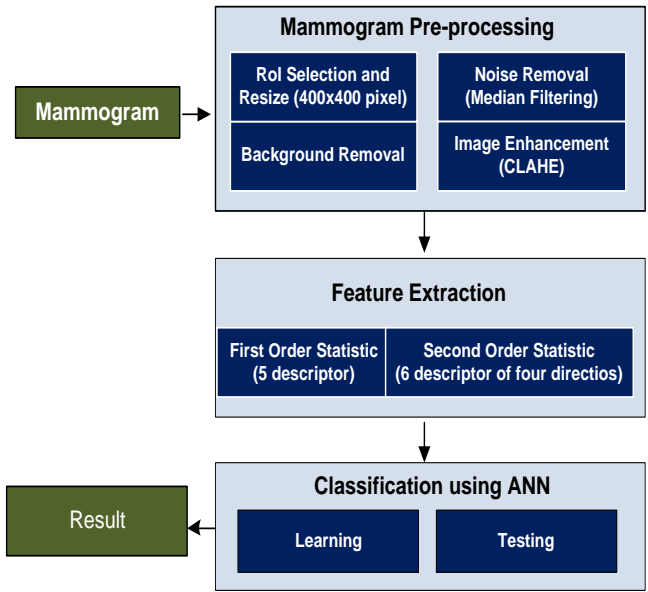

Figure 1. the proposed model

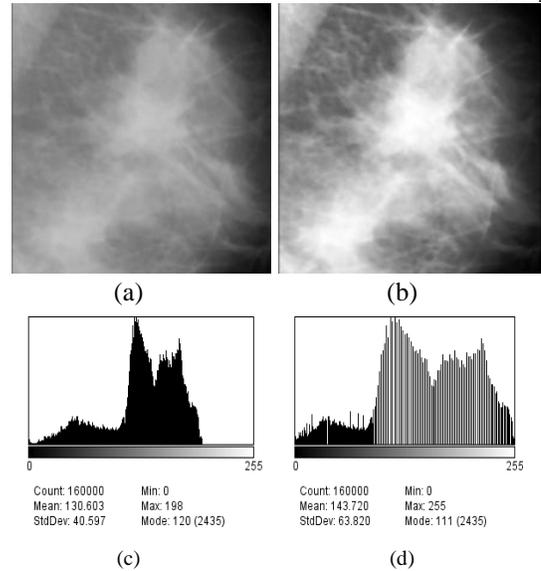

Figure 3. (a) an image median filtering results, (b) image operating results CLAHE), and (c-d) their histogram

## B. Methodology

### 1) Pre-processing

The preprocessing was carried out to improve the quality of the image of mammogram before feature extraction. There are several processes that are performed at this stage : cropping on the Region of Interest (RoI), resizing an image of a mammogram to be (400 x 400 pixel), removing background, reducing noise with median filtering, improving the contrast of the image by CLAHE method (Contrast-Limited Adaptive Histogram Equalization) [9]. The results of each stage of their histogram are shown in Figure 2 and 3.

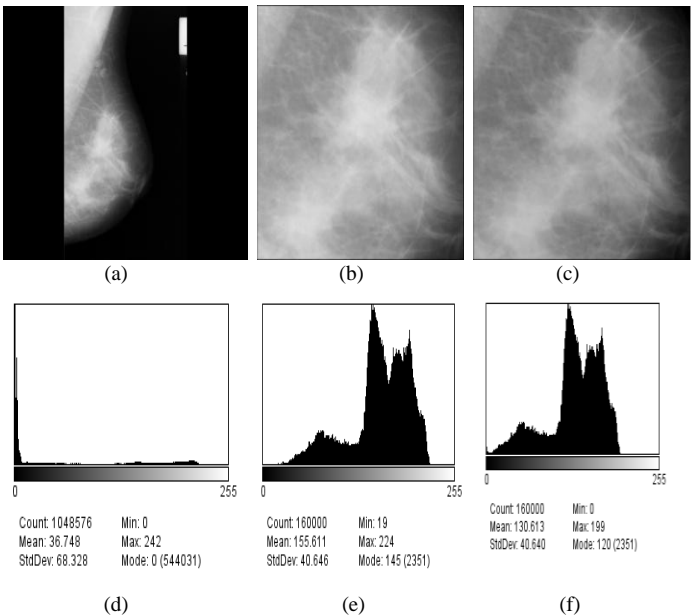

Figure 2. (a) an image median filtering results, (b) image operating results CLAHE), and (c-d) their histogram

### 2) Feature extraction

The difference in mass between benign and malignant on the image of a mammogram can be distinguished from their textures. Feature extraction is the first step in performing the classification and interpretation of images. The statistical feature extraction of statistical parameter of the image of interest. There are five parameters being extracted for the first order. In addition, variance parameter is extracted for the second order.

#### a) The first order statistics

First order feature extraction is a method of retrieval based on characteristics of the image histogram. The Histogram shows the probability of occurrence of the value of the degree of grayscale pixels in an image. From the values produced in the histogram can be calculated several parameters of the first order namely : mean, variance, skewness, kurtosis and entropy.

- Mean

  It shows the size of the dispersion of an image.

  $$\mu = \sum_n f_n P(f_n) \qquad (1)$$

- Variance

  Variance shows the variations of the element on histogram of an image.

  $$\sigma^2 = \sum_n (f_n - \mu)^2 p(f_n) \qquad (2)$$

- Skewness

  It indicates the relative level of the slope of the curve on the histogram of an image.

  $$\alpha_3 = \frac{1}{\sigma^3} \sum_n (f_n - \mu)^3 p(f_n) \qquad (3)$$

- Kurtosis

  It indicates the level of sharpness relatively curve on the histogram of an image.

  $$\alpha_4 = \frac{1}{\sigma^4} \sum_n (f_n - \mu)^4 p(f_n) - 3 \quad (4)$$

- Entropy

  Entropy shows the size of the irregular shape of an image

  $$H = - \sum_n p(f_n)^2 \log p(f_n) \quad (5)$$

  *b) The second order statistics*

  One of the techniques to obtain the second order features is calculating the probability of a relationship between two pixels at a distance and orientation invariant. There are several stages for the second order, the first is forming of the matrix co-occurrence and the second is specifiying the characteristics as a function of the matrix. Co-occurrence is the value of a pixel's neighbors in the the distance (d) and orientation angle (θ). A unit of distance is used in pixels and orientation in degree. Orientation is formed at four directions with angular interval angle of 45° namely 0°, 45°, 90°, and 135°. The distance between pixels is usually equal to one pixel. Haralick et al [10] propose various types of texture features that can be extracted from the matrix co-occurrence. This work uses 6 features of the second order statistics i.e. Angular Second Moment, Contrast, Correlation, Variance, Inverse Difference Moment and Entropy. P is defined by [11] :

- Entropy

  Entropy shows the randomness of the pixels of an image .The higher entropy value, the more random texture.

  $$\text{Entropy} = - \sum_{i,j} P(i,j) \log P(i,j) \quad (6)$$

- Contrast

  Contrast shows the local variation in image content. The higher the contrast, the higher the level of contrast.

  $$\text{Contrast} = \sum_{i,j} |i-j|^2 P(i,j) \quad (7)$$

- Correlation

  Correlation indicates the size of the linear relationship of the neighborhood pixel gray level.

  $$\text{Correlation} = \sum_{i,j} \frac{(i-\mu_i)(j-\mu_j)P(i,j)}{\sigma_i \sigma_j} \quad (8)$$

- Angular Second Moment (ASM)

  ASM shows the homogeneity properties of an image size or the size of the proximity of each element of the occurrence matrix.

  $$\text{ASM} = \sum_{i,j} \frac{P(i,j)}{1+|i-j|} \quad (9)$$

- Inverse Difference Moment (IDM)

  IDM is the opposite of contrast .The higher the value of IDM, the lower the level of contrast .

  $$\text{IDM} = \sum_{i,j} \frac{P(i,j)^2}{|i-j|} \quad (10)$$

- Variance

  Variance shows the variations of the matrix co-occurrence elements.

  $$Variance = - \sum_i \sum_j p(i,j)^2 \log p(i,j) \quad (11)$$

*3) Classification*

The process of learning for this classification uses backpropagation learning with the architecture of multi-layer perceptron. Backpropagation is a type of artificial neural network (ANN) learning method which most widely used and have a good performance. The difference with the perceptron, is that the backpropagation learning method has many layers (multilayer), its layer may have different activation function. The backpropagation has also more powerful learning ability [12]. There are many parameters that must be specified before the training is carried out, i.e. the number of hidden layer, the number of neurons in the hidden layer, activation function, the learning rate and the conditions that stop learning. Related to the number of neurons in the hidden layer there is no certainty about how much the most optimal number of nodes. In neural network, the number of nodes depends on the pattern of any dataset's uniqueness. Therefore the number of nodes in the hidden layer can be calculated using equations 12 and 13.

$$\text{Hidden Unit} = (n+1) * \frac{2}{3} \quad (12)$$

where n is the number of nodes in the input layer (rounding down) [13].

$$N_h = \overline{N_i * N_o} \quad (13)$$

where $N_h$ is the number of neurons in hidden layer, $N_i$ is the number of nodes in input layer and $N_o$ is the neuron in output layer (rounding up) [14]. As for the learning rate = 0.3, error goal = 1e-4, momentum = 0.9 and sigmoid activation function is used. The sigmoid bipolar function is the most commonly function used. Usually, the sigmoid bipolar is the commonly used fot the backpropagation training method.

In this stage of experiment using digital mammogram images of 50 images, 80% (40 of 50) for training and the rest for testing. After the feature extraction is carried out, the result then are classified into nine texture descriptors. The nine texture descriptors are *(1)* five descriptors are extracted using the first order statistics extract (mean, variance, skewness, kurtosis and entropy); *(2)* six descriptors are extracted using the average of second order statistics extract with details of six texture descriptors = (4 x 1 distance x 6 descriptors)/4; *(3)* twenty four descriptors are extracted using the second order statistics for each direction with details (4 direction x 1 distance x 6 descriptors); *(4-7)* five descriptors are extracted using the second order statistics. They have the same as the number of descriptors, but they have different directions ($0^0, 45^0, 90^0, 135^0$); *(8)* eleven descriptors are extracted using the first (5 descriptors) and the average of the second (6 descriptors) order statistics for four directions *(9)* twenty nine descriptors are extracted using the first (5 descriptors) and second order statisctics for four directions (4 direction x 1 distance x 6 descriptor). The nine descriptors are then inputed to the ANN with the number nodes in the hidden layer is calculated using formulas 12 and 13. The ANN with hidden nodes calculated using formula 12 is called MLP-1, while the other calculated using formula 13.

The architecture of MLP uses here is M-N-O, where M, N, O are the number of nodes in input layer, hidden layer and output layer respectively. For example the architecture of 5-4-1 menas that it has 5 nodes in input layer, 4 nodes in hidden layer and one node in output layer such as shown in the row two and column three and four in the table 1.

TABLE I. THE NUMBER OF NODE ON HIDDEN LAYER FOR EACH TEXTURE DESCRIPTOR IN MLP-1 AND MLP-2.

| No | Texture Descriptor | Input Unit | Hidden Unit | | Output Unit |
|----|--------------------|------------|-------------|-------|-------------|
|    |                    |            | *MLP-1*     | *MLP-2* |           |
| 1  | first order        | 5          | 4           | 3     | 1           |
| 2  | second order (mean) | 6         | 4           | 3     | 1           |
| 3  | second order (combination) | 24 | 16         | 5     | 1           |
| 4  | second order- $0^0$ | 6         | 4           | 3     | 1           |
| 5  | second order- $45^0$ | 6        | 4           | 3     | 1           |
| 6  | second order-$90^0$ | 6         | 4           | 3     | 1           |
| 7  | second order-$135^0$ | 6        | 4           | 3     | 1           |
| 8  | first&second order (mean) | 11 | 8           | 4     | 1           |
| 9  | first&second order (combination) | 29 | 20   | 6     | 1           |

### III. RESULT AND DISCUSSION

There are three stages of the processes carried out in this research are pre-processing, feature extraction and classification. The results of the training and testing for the MLP-1 in classification precess for the MLP-1 having the regression value 1 are second order (mean), second order (combination) and second order direction ($0^0$ and $135^0$). While for the MLP-2 are second order (mean), second order (combination), second order for all directions ($0^0, 45^0, 90^0, 135^0$), first and second order (mean) and first and second order (combination). X axis represent the texture descriptor used, for example the value 1 of X axis means "first order" used such as shown in table 1 column 2 row 2. The figure 4 shows that the best value for texture descriptors uses here are second order (mean), second order (combination) and second order with direction ($0^0$ and $135^0$), in this figure 4 is shown number (2, 3, 4 and 5) on the X axis. These values of descriptors have regression values are 1.

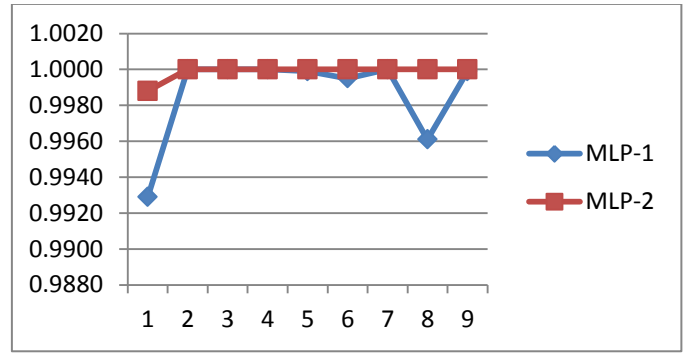

Figure 4. The value of regression for MLP-1 and MLP-2

The number epoches for the MLP-1 and MKP-2 are graphically shown in figure 5. The better architecture is that has the smaller number of epoches. In this research the number of epoches assumed to be good is less than 1000 epoches. The figure 5 shows that there are three texture descriptors having the number of epoches for the MLP-1 398, 884 and 102 consecutively (3,4 and 9). While the other there are four texture descriptors having the number of epoches 110, 192, 892 and 365 consecutively (3, 6, 8 and 9). The figure 5 shows that the best value for texture descriptors uses here are second order (combination) and the first + second order (combination), in this figure 5 is shown number 3 and 9 on the X axis. These values of descriptors have the number of epoches are (398 and 102 for the MLP-1) and (110 and 365 for the MLP-2).

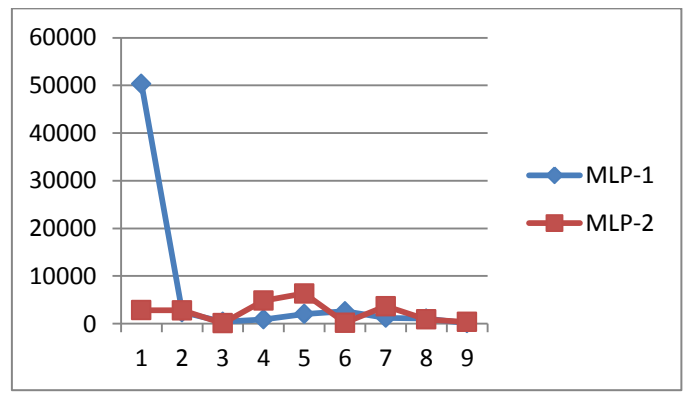

Figure 5. The number of epoches for MLP-1 and MLP-2

## IV. CONCLUSIONS

The experimentas results show that having two types of classification carried out using the regression method and considering the less number of epoches. The best texture descriptor as selected when the value of regression 1 appears in both the MLP-1 and the MLP-2 with the number of epoches less than 1000. In this case the best selected texture descriptor is the second order (combination) using all direction ($0^0, 45^0, 90^0, 135^0$) that have twenty four descriptors.

### ACKNOWLEDGMENT

The authors would like to thank Sunan Kalijaga State Islamic University (http://www.uin-suka.ac.id) for funding the research, and the Department of Computer Science and Electronics Gadjah Mada University (http://mkom.ugm.ac.id) for providing technical support for the research

### AUTHORS PROFILE

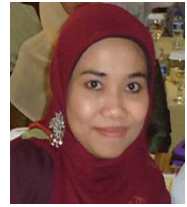

**Shofwatul 'Uyun** is a Full Time Lecturer at the department of Informatics, Faculty of Science and Technology, Sunan Kalijaga State Islamic University (UIN) in Yogyakarta, Indonesia. She is currently taking her Doctoral Program at the Department of Computer Science and Electronics, Gadjah Mada University in Yogyakarta, Indonesia

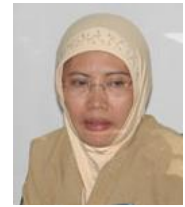

**Sri Hartati** is an Associate Professor and head of graduate program of Computer Science at the Department of Computer Science and Electronics, Gadjah Mada University in Yogyakarta, Indonesia. She obtained her Bachelor degree in Electronics and Instrumentation from the Gadjah Mada University. She received her M.Sc. and PhD in Computer Science from the University of New Brunswick, Canada. Her research interests are artificial intelligence and decision support system.

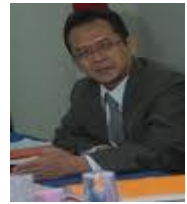

**Agus Harjoko** is an Associate Professor at the Department of Computer Science and Electronics, Gadjah Mada University in Yogyakarta, Indonesia. He obtained his Bachelor degree in Electronics and Instrumentation from the Gadjah Mada University. He received his M.Sc. and PhD in Computer Science from the University of New Brunswick, Canada. His research interests are image processing and pattern recognition.

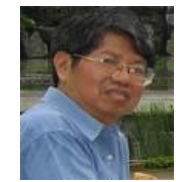

**Subanar** is a Professor at the Department of Mathematics, Gadjah Mada University in Yogyakarta, Indonesia. He was graduated as Bachelor of Mathematics from Gadjah Mada University and Ph.D (Statistics) at University of Wisconsin, Madison, USA.